\documentclass{article}

\PassOptionsToPackage{numbers,sort&compress}{natbib}
\usepackage[preprint]{neurips_2026}

\usepackage[utf8]{inputenc}
\usepackage[T1]{fontenc}
\usepackage{booktabs}
\usepackage{graphicx}
\usepackage{float}
\usepackage{amsmath}
\usepackage{amssymb}
\usepackage{microtype}
\usepackage{xcolor}
\usepackage{hyperref}
\usepackage{url}
\usepackage{needspace}

\graphicspath{{figures/}}

\title{Request Order Matters: Cache-History Sensitivity in Selective KV-Cache Reuse for Rolling Agents}

\author{%
  Tiffany Gu\thanks{Equal contribution.} \\
  Carnegie Mellon University \\
  \texttt{tg2@andrew.cmu.edu} \\
  \And
  Annie Guan\footnotemark[1] \\
  Capital One \\
  \texttt{xin.guan2@capitalone.com} \\
  \AND
  Manshu Huang \\
  University of Chicago \\
  \texttt{huangms@uchicago.edu} \\
  \And
  Nitin Rao \\
  University of California, Berkeley \\
  \texttt{nitinrao@berkeley.edu} \\
  \And
  Siddhant Shah \\
  University of Southern California \\
  \texttt{shahsd@usc.edu} \\
  \And
  Margaret Capetz \\
  University of Washington \\
  \texttt{mcapetz@uw.edu} \\
}

\begin{document}
\raggedbottom

\maketitle

\begin{abstract}
Long-running agents repeatedly call an LLM while retaining most of their document window, evicting old documents, and appending new ones. These rolling updates break exact prefix caching and motivate non-prefix KV-cache reuse with selective recomputation. We show that persistent KV-cache reuse with selective recomputation can be history-dependent: in our rolling-agent workload, an unchanged prompt can produce different answers depending on the requests processed before it. At a matched 5\% recomputation budget, document-aligned recomputation reduces answer variation across request orders from 69.0\% with CacheBlend's token top-$k$ policy to 26.1\%. When each prompt is evaluated after a different sequence of preceding requests, document-aligned recomputation improves fidelity to full prefill by 34.5--52.5 percentage points over token top-$k$, while both policies achieve approximately 5.7$\times$ median TTFT speedup. Our ablation study shows that, in our rolling-agent workload, contiguity is the main factor associated with robust selective recomputation.
\end{abstract}

\section{Introduction}

Long-running agents repeatedly call an LLM while retaining information from previous steps. A research agent may keep most papers while replacing outdated evidence; a financial monitoring agent may discard one stale quote or filing and append a new headline, price update, or 8-K title. Recomputing the entire prompt after every small update wastes prefill computation and increases time to first token.

Exact prefix caching avoids repeated prefill only when a new request begins with the same token sequence as a cached request. A rolling update breaks this condition. Once the oldest document is removed, the retained documents move to earlier positions and acquire different preceding context. Their text is unchanged, but their previously computed KV states are no longer the exact states produced by a full prefill of the new prompt.

CacheBlend addresses this problem by loading cached KV states for reusable chunks and selectively recomputing tokens whose current keys deviate most from their cached keys~\citep{cacheblend}. LMCache implements this idea with token top-$k$: for a fixed recomputation budget, it selects the individual tokens with the largest deviations~\citep{lmcache}. Because tokens are ranked independently, the resulting repair can be scattered across hundreds of disjoint regions.

We study two questions. First, when the number of recomputed tokens is fixed, does selecting a few contiguous spans work better than selecting scattered tokens with the largest KV deviations? Second, when KV cache state persists across requests, can different sequences of earlier requests cause the same current prompt to produce different answers? This is important for agents because each LLM call inherits cache state created by previous calls.

\begin{figure}[t]
  \centering
  \includegraphics[width=\linewidth]{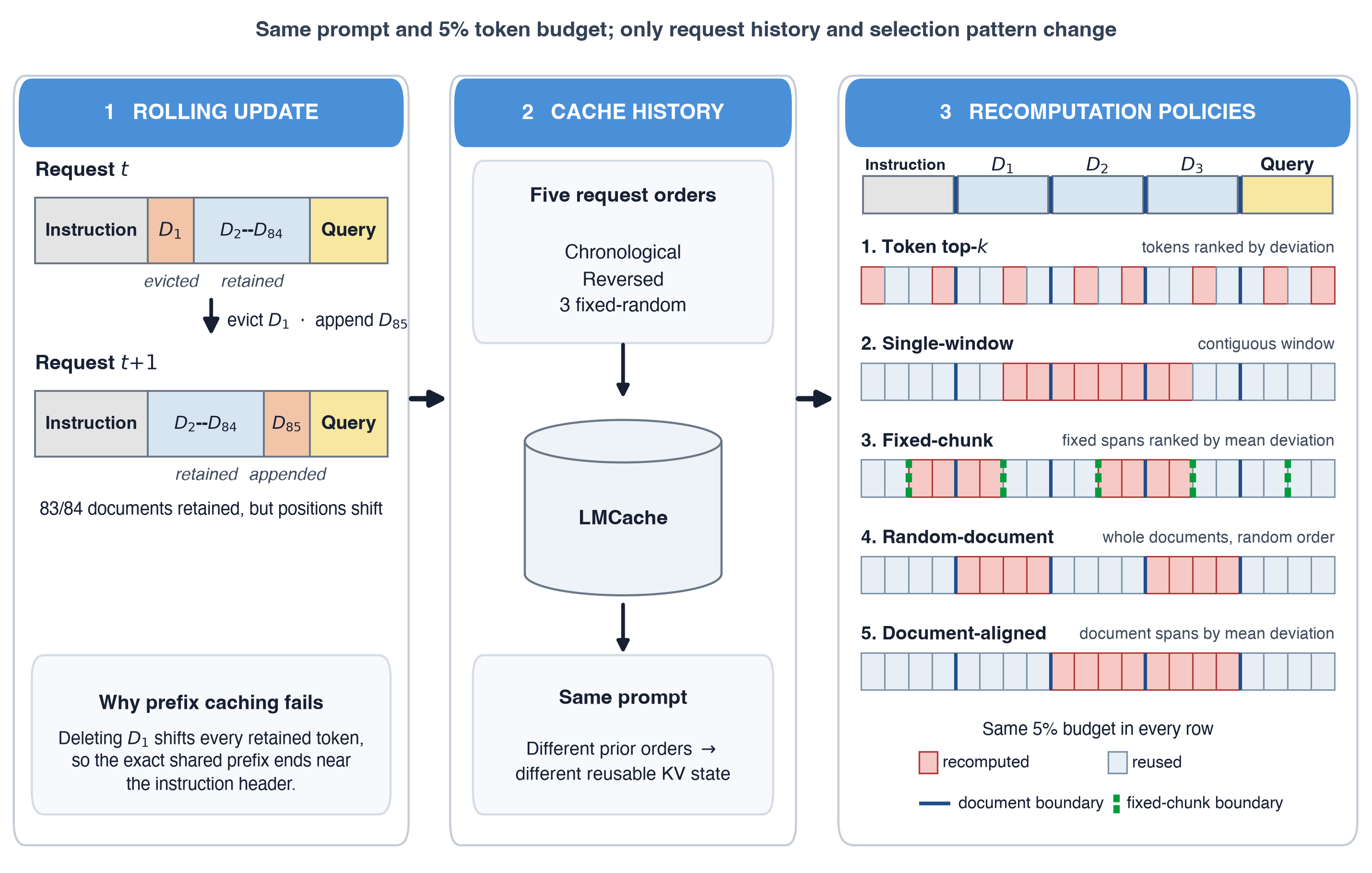}
  \caption{Overview of the rolling workload, persistent cache history, and equal-budget recomputation policies. Different request orders expose the same prompt to different cached chunk KV states; every policy recomputes 5\% of reusable tokens.}
  \label{fig:overview}
\end{figure}

We compare token top-$k$ with document-aligned recomputation at the same 5\% budget. An ablation additionally tests a single contiguous window, fixed chunks crossing document boundaries, and randomly ranked document spans. This design separates contiguity from document boundaries and deviation-based ranking. Figure~\ref{fig:overview} summarizes the rolling workload, persistent cache histories, and five equal-budget recomputation policies.

Several systems have shown that non-prefix KV-cache reuse can reduce prefill latency~\citep{cacheblend,epic,kvlink,cacheclip,minipic,lmcache}. We keep both policies at the same recomputation count and measure time to first token on the same rolling requests, so the comparison holds the prefill budget fixed rather than trading extra recomputation for higher fidelity.

We make two contributions:
\begin{enumerate}
  \item We show that selective KV-cache reuse depends on request history. The same prompts produce substantially more answer variation with token top-$k$ than with document-aligned recomputation when request order changes.
  \item We show that recomputing a few contiguous regions is more robust than recomputing scattered high-deviation tokens under the tested persistent cache history.
\end{enumerate}

\section{Background}

\subsection{Prefill latency in rolling document workloads}

LLM inference begins with a prefill stage that processes the input prompt and creates per-layer key and value states for its tokens. Longer prompts increase prefill work and time to first token. Prefix caching reduces this cost by reusing KV states for the longest identical token prefix shared with an earlier request~\citep{pagedattention,sglang}. Reusing an exact prefix preserves the exact model computation for the reused tokens, but deleting or inserting content near the beginning prevents reuse for all unchanged text that follows.

Rolling document workloads expose this limitation directly. Request $t$ contains a fixed instruction, documents $D_1,\ldots,D_{84}$, and a final query. Request $t+1$ evicts $D_1$, retains $D_2,\ldots,D_{84}$, and appends $D_{85}$. Although 83 of the 84 documents persist, every retained document shifts to an earlier position and acquires different preceding context, as illustrated in Figure~\ref{fig:overview}. The longest shared prefix is therefore little more than the fixed instruction, and standard prefix caching cannot reuse any of the retained documents' prior KV states.

A non-prefix reuse system instead stores the KV states of individual chunks and loads them when those chunks appear in later prompts. Simply combining independently computed chunk caches is fast but approximate: each chunk was processed without the preceding chunks in the current prompt and therefore lacks the cross-chunk attention that a full prefill would create. CacheBlend mitigates this inconsistency by selectively recomputing a subset of tokens under the current prompt~\citep{cacheblend}. The retrieved chunks determine which cached KV states are loaded, while the recomputation selector determines which tokens within those chunks are recomputed. Related approaches modify the selection signal, linking mechanism, position handling, attention structure, or model training~\citep{epic,kvlink,prophetkv,cacheclip,minipic,chunkstudy}.

\subsection{CacheBlend's token-level selective recomputation}

After reusable KV states are loaded, LMCache's CacheBlend implementation processes the prompt layer by layer. At a configured selection layer, it compares each reusable token's current key $K_i^{\mathrm{current}}$ with its loaded key $K_i^{\mathrm{cached}}$:
\begin{equation}
  d_i=\lVert K_i^{\mathrm{current}}-K_i^{\mathrm{cached}}\rVert_2^2.
  \label{eq:deviation}
\end{equation}
Here, $d_i$ is token $i$'s deviation score. A larger value means that the token's loaded key differs more from the key computed under the current prompt, so CacheBlend treats that token as a higher priority for recomputation.

For $N$ reusable tokens and recomputation ratio $r$, LMCache selects
\begin{equation}
  k=\max(1,\lfloor rN \rfloor)
  \label{eq:budget}
\end{equation}
tokens. Its default policy selects the $k$ tokens with the largest $d_i$ values. The selector runs once at the selection layer, and the same selected token positions are recomputed at every subsequent layer without being reranked. Selected positions use KV states computed under the current prompt, while unselected positions retain their loaded KV states. This policy captures the largest individual deviations but can scatter recomputation across many regions.

\subsection{Span-level selective recomputation}

Span-level selective recomputation allocates the recomputation budget to contiguous ranges of reusable tokens rather than isolated token positions. We consider four span-based policies that use the same reusable tokens and recompute exactly $k$ tokens. For a candidate span $S_j$, we define its average KV-deviation score as
\begin{equation}
  \rho_j=\frac{1}{|S_j|}\sum_{i\in S_j}d_i,
  \label{eq:span-score}
\end{equation}
where $d_i$ is the per-token deviation from~\eqref{eq:deviation}.
This score measures how much the tokens in a span differ from their cached states on average. Normalizing by span length prevents longer spans from receiving a higher score merely because they contain more tokens. The four policies differ in how they define and rank their candidate spans:

\begin{itemize}
  \item \textbf{Single-window} selects the contiguous $k$-token window with the greatest total deviation, $\sum_i d_i$. Because every candidate window contains $k$ tokens, ranking by total or average KV deviation gives the same result.
  \item \textbf{Fixed-chunk} divides the reusable prompt into equal-size spans that cross document boundaries and prioritizes the spans with the highest average KV-deviation scores $\rho_j$.
  \item \textbf{Random-document} uses document spans but ranks them in a deterministic random order, ignoring the deviation scores.
  \item \textbf{Document-aligned} uses the same document spans and ranks them by average KV-deviation score $\rho_j$.
\end{itemize}

We use document-aligned recomputation as the span-based policy in the cache-history sensitivity experiment and TTFT benchmark. The other three span policies serve as ablation controls that isolate the effects of contiguity, document boundaries, and deviation-based ranking. Only random-document and document-aligned use document boundaries; single-window and fixed-chunk define contiguous regions independently of those boundaries.

Our document-aligned policy divides the reusable prompt into the instruction span, individual document spans, and the final query when it is reusable. It ranks spans by $\rho_j$ and selects complete spans while they fit within the $k$-token budget. If tokens remain, it selects the contiguous window with the greatest total deviation from the highest-ranked remaining span. The final selection contains exactly $k$ tokens. Document boundaries provide a practical way to define contiguous regions because they are already available in document-oriented prompts and require neither a learned segmenter nor a manually chosen chunk size. Figure~\ref{fig:overview} summarizes how the five policies place the same recomputation budget.

\section{Related Work}

\paragraph{Selective recomputation for non-prefix reuse.}
CacheBlend introduces selective recomputation of high-deviation tokens to repair cross-document interactions after loading independently cached KV states~\citep{cacheblend}. ProphetKV makes selection query-driven and combines signals across layers~\citep{prophetkv}. CacheClip uses an auxiliary model to identify important tokens and groups nearby selections to preserve local coherence~\citep{cacheclip}. Unlike CacheClip, which incorporates grouping as one component of a complete method, we compare scattered and span-based selection at the same realized recomputation count and isolate the effect of selection pattern. A recent experimental study compares recomputation and attention-reshaping strategies for chunk-level caching and similarly emphasizes that no single repair mechanism is sufficient across settings~\citep{chunkstudy}. These methods study repair quality for assembled prompts; we instead ask whether a repair policy remains reliable as persistent cache state accumulates across rolling requests.

\paragraph{Position-independent and document-level reuse.}
EPIC formalizes position-independent caching and introduces LegoLink, which recomputes a small fixed set of initial tokens from each reusable chunk to suppress spurious attention sinks~\citep{epic}. This is contiguous rather than scattered selection, though it spreads many small spans across all chunks rather than concentrating the budget as our policies do. KVLink adjusts positional encodings and inserts trainable link tokens between independently encoded documents~\citep{kvlink}. MiniPIC exposes position-free KV storage and span-oriented reuse primitives inside vLLM~\citep{minipic}. These systems expand reuse beyond exact prefixes; our results identify persistent request history as an additional evaluation dimension.

\paragraph{Prefix caching and KV-cache management.}
vLLM introduces PagedAttention, which manages KV-cache memory in noncontiguous blocks to reduce fragmentation and enable efficient sharing~\citep{pagedattention}. SGLang's RadixAttention organizes reusable prefixes in an LRU-managed radix tree and combines prefix matching, eviction, and cache-aware scheduling~\citep{sglang}. LMCache externalizes KV storage and movement across engines and memory tiers and provides the CacheBlend implementation used in our experiments~\citep{lmcache}. Together, these systems make KV state a persistent serving resource. Our work studies selective repair when rolling prompts reuse content beyond a stable prefix and persistent history changes the cached state.

\section{Evaluation and Results}

Our experiments ask four questions: Does request order affect the outputs of selective KV reuse? Is document-aligned recomputation less sensitive to request order than token top-$k$? How much does selective reuse reduce TTFT? Which properties of span selection explain the observed difference?

\subsection{Experimental Setup}

\paragraph{System configuration.}
We use Qwen3-8B with greedy decoding (temperature 0) and a fixed random seed, LMCache 0.5.1~\citep{lmcache}, and vLLM 0.23.0~\citep{pagedattention}. The vLLM prefix cache is disabled so that the experiment measures LMCache's non-prefix reuse path. Token top-$k$ and document-aligned recomputation use the same model, prompts, reusable-token interface, deviation metric, and 5\% recomputation rule.

\paragraph{Dataset and rolling agent workload.}
The full event dataset contains 8,995 real records for 15 U.S. equities: 7,515 Yahoo Finance daily price records~\citep{yahoo}, 792 FNSPID news headlines~\citep{fnspid}, and 688 SEC EDGAR filing titles~\citep{edgar}. News and filing dates are mapped onto each equity's price calendar to interleave the three channels while preserving order within each channel.

The workload models an agent with a bounded document window. Before each LLM call, the agent evicts one old document, retains 83 documents, and appends one new document. For example, the window $[D_1,\ldots,D_{84}]$ becomes $[D_2,\ldots,D_{85}]$ at the next call. The event stream produces 7,735 rolling updates with a fixed window of 84 documents and stride 1. The final query identifies one observation in the current window and asks the model to copy the requested fields from that observation verbatim. This deterministic extraction task supports exact answer comparison. Prompts contain approximately 7.2K--7.7K tokens.

\paragraph{Evaluation protocol.}
We process the same requests in five orders. The prompt for each request remains unchanged; only the order in which complete requests enter the serving system changes. This varies the persistent cache state created by earlier requests and allows us to test whether the same prompt remains stable under different cache histories:
\begin{itemize}
  \item \textbf{Chronological:} models a rolling stream processed normally in source order.
  \item \textbf{Fixed-random:} uses three predetermined shuffled orders to represent request interleavings when a shared cache serves multiple agents.
  \item \textbf{Reversed:} processes the source stream in the opposite direction as a deterministic stress test.
\end{itemize}

Each combination of recomputation policy and request order starts with empty LMCache state, retains cache state across requests, and shares no state with other runs. For each rolling update, we first populate the previous window's document KV caches and then evaluate the updated window.

We report two cache-history sensitivity metrics. For each request, we compare the answers produced under all ten pairs of the five request orders. Answer variation is the percentage of these comparisons in which the answers differ. Fidelity is exact agreement with the full-prefill answer. We also compare request-level TTFT for full prefill, token top-$k$, and document-aligned recomputation.

\subsection{Cache History Sensitivity}

Request order changes the answer. Answer variation is 69.0\% for token top-$k$ and 26.1\% for document-aligned recomputation. Requests are issued one at a time to rule out batch-composition effects, and an independent repeat of the chronological order reproduces every answer exactly, ruling out run-to-run variation.

\begin{figure}[H]
  \centering
  \includegraphics[width=\linewidth]{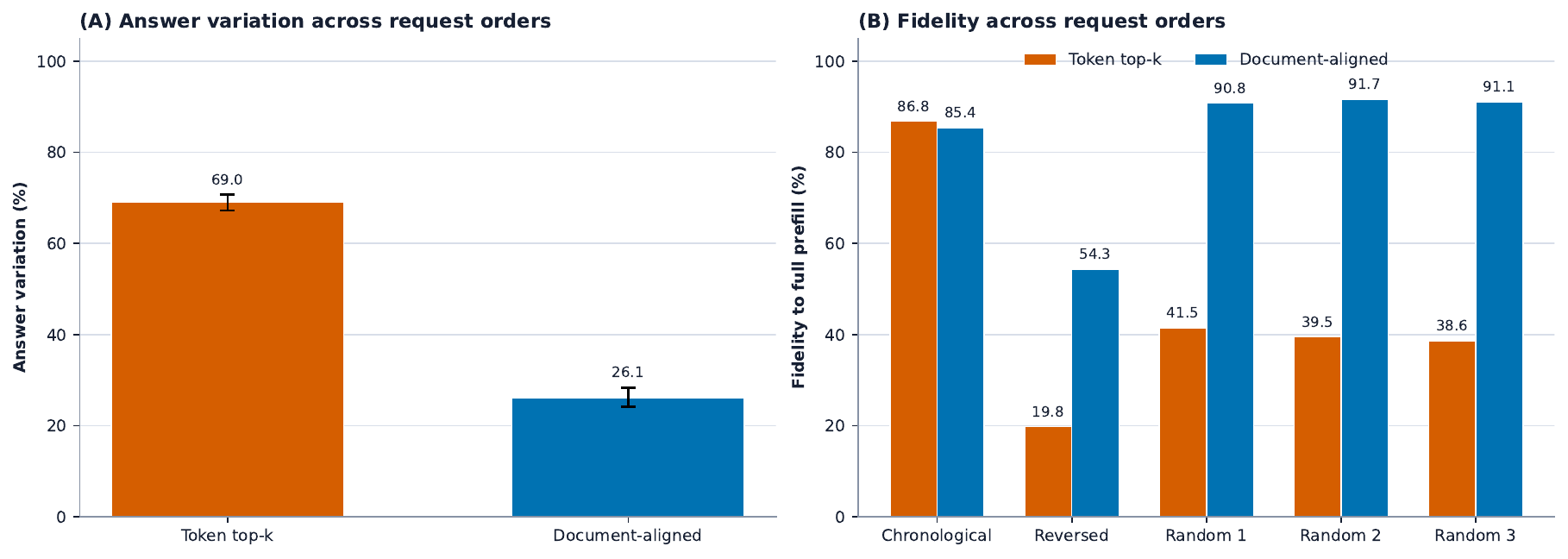}
  \caption{(A) Answer variation, measured as mean pairwise exact-answer disagreement across the five request orders; error bars show 95\% confidence intervals. (B) Fidelity to full prefill in each request order. Both recomputation policies use a 5\% recomputation ratio.}
  \label{fig:experimental-results}
\end{figure}

Request order also affects fidelity. Under chronological order, we detect no statistically significant difference: 86.8\% for token top-$k$ and 85.4\% for document-aligned recomputation ($p=0.12$). Under the other four orders, fidelity ranges from 19.8--41.5\% for token top-$k$ and 54.3--91.7\% for document-aligned recomputation (all $p<0.0001$). Document-aligned recomputation is more robust but not immune: its fidelity falls to 54.3\% under the reversed-order stress test (Figure~\ref{fig:experimental-results}).

In a 500-request control that resets the cache before each request, we detect no statistically significant difference in fidelity between token top-$k$ and document-aligned recomputation (94.0\% vs.\ 92.6\%, $p=0.23$). This control suggests that the advantage of document-aligned recomputation depends on cache state accumulated across requests: when that history is removed, the two policies perform similarly.

\subsection{Time to First Token}

Table~\ref{tab:ttft} reports request-level median and P95 TTFT and median speedup over full prefill on an NVIDIA L4. Both selective-reuse policies achieve approximately 5.7$\times$ median speedup and have similar latency.

\begin{table}[H]
  \caption{Request-level time to first token.}
  \label{tab:ttft}
  \centering
  \small
  \begin{tabular}{lrrr}
    \toprule
    Policy & Median TTFT & P95 TTFT & \shortstack{Median speedup\\vs.\ full prefill} \\
    \midrule
    Full prefill & 2,408~ms & 2,524~ms & 1.00$\times$ \\
    Token top-$k$, 5\% & 426~ms & 459~ms & 5.68$\times$ \\
    Document-aligned, 5\% & 415~ms & 477~ms & 5.74$\times$ \\
    \bottomrule
  \end{tabular}
\end{table}

\subsection{Selection-pattern ablation}

To identify which property of document-aligned recomputation drives its behavior, we compare it with three span-based controls and token top-$k$. The ablation evaluates all five policies on the same 500-request cohort under chronological and fixed-random request orders, using a 5\% recomputation ratio and matching the exact number of recomputed tokens. Token top-$k$ selects scattered high-deviation tokens. Single-window selects one continuous region. Fixed-chunk partitions the reusable prompt into 92-token spans, matching the cohort's median document-span length, offsets them so they cross document boundaries, and prioritizes the spans with the highest average KV-deviation scores. Random-document uses document spans but ignores deviation ranking. Document-aligned ranks document spans by the same average KV-deviation score (see Figure~\ref{fig:overview}).

\begin{figure}[H]
  \centering
  \includegraphics[width=0.78\linewidth]{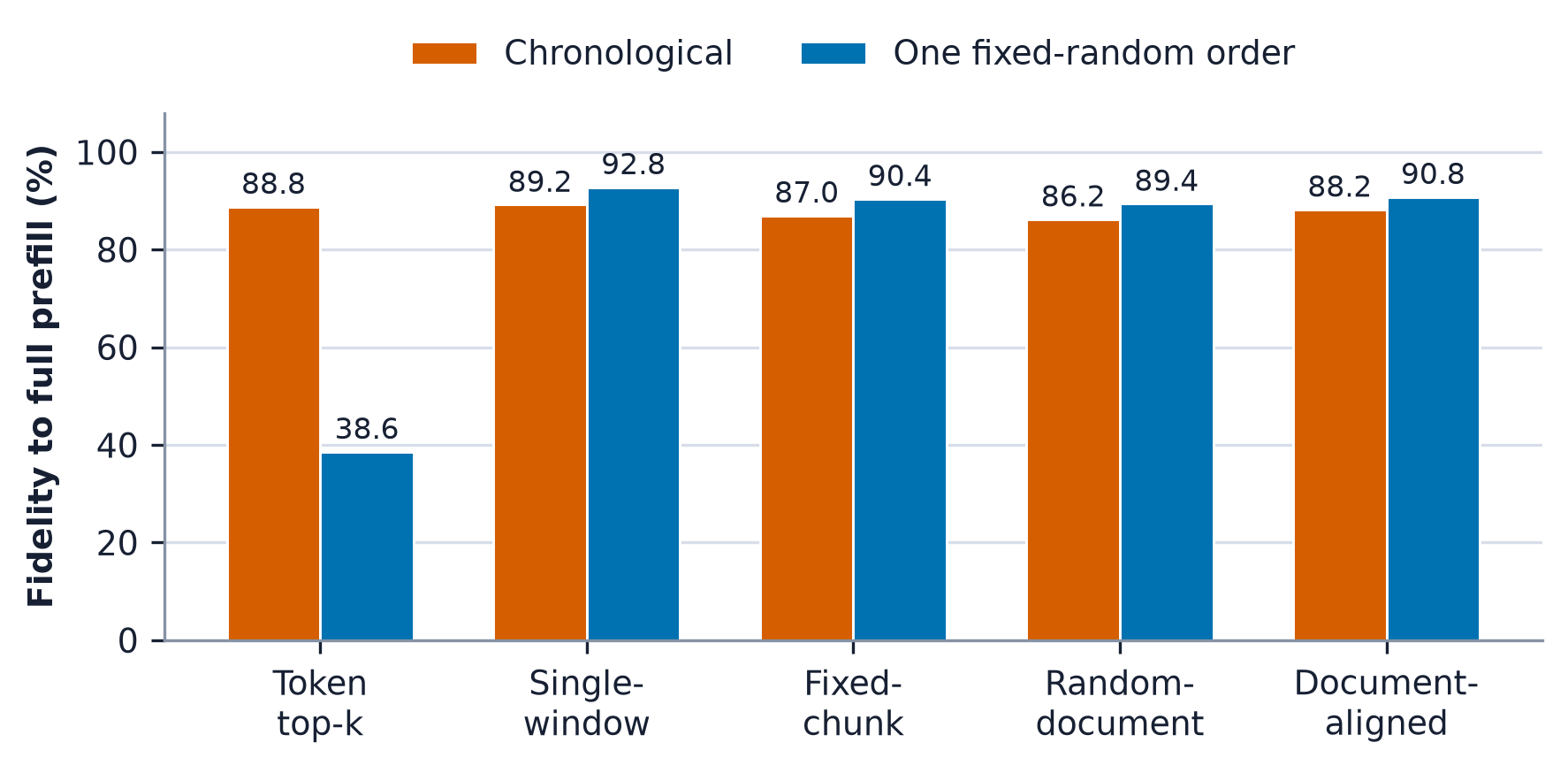}
  \caption{Fidelity of five equal-budget selection policies on the same prompts under chronological and one fixed-random order. Every policy uses a 5\% recomputation ratio and matches the realized token count within each order. Error bars show 95\% Wilson score intervals.}
  \label{fig:ablation}
\end{figure}

All five policies recompute exactly the same total number of tokens within each order: 188,562 in chronological execution and 188,721 in the fixed-random order. The median per-request count is 386 and 387 tokens, respectively. Because every policy recomputes the same number of tokens, the fidelity differences come from where those tokens are selected, not from doing more recomputation (Figure~\ref{fig:ablation}).

Three fidelity patterns emerge:
\begin{itemize}
  \item \textbf{Similar fidelity in chronological order.} All five policies achieve 86.2--89.2\% fidelity, and no comparison supports a clear advantage for any policy.
  \item \textbf{Span policies retain high fidelity under the fixed-random order.} Token top-$k$ falls to 38.6\% fidelity, while the four span policies achieve 89.4--92.8\%. Each span policy improves fidelity over token top-$k$ by 50.8--54.2 percentage points (all $p<0.0001$).
  \item \textbf{Document boundaries and deviation-based span ranking show no statistically significant additional advantage.} Single-window achieves the highest fixed-random fidelity at 92.8\%. Document-aligned does not differ significantly from single-window ($p=0.076$), fixed-chunk ($p=0.85$), or random-document selection ($p=0.37$).
\end{itemize}

Token top-$k$ is designed to capture the largest individual KV deviations, yet it performs worst under the fixed-random order. Table~\ref{tab:geometry} reports two properties of each selection: how many separate contiguous regions receive the budget (segments), and what share of the total selection-layer KV-deviation score those tokens carry (captured deviation).

\begin{table}[H]
  \caption{Selection fragmentation and captured deviation. Each cell reports chronological / fixed-random results.}
  \label{tab:geometry}
  \centering
  \small
  \begin{tabular}{lcc}
    \toprule
    Policy & Median segments & Median captured deviation \\
    \midrule
    Token top-$k$ & 199 / 180.5 & 98.8\% / 99.7\% \\
    Single-window & 1 / 1 & 7.2\% / 6.1\% \\
    Fixed-chunk & 3 / 5 & 6.9\% / 7.1\% \\
    Random-document & 6 / 6 & 4.8\% / 4.8\% \\
    Document-aligned & 3 / 6 & 7.2\% / 6.9\% \\
    \bottomrule
  \end{tabular}
\end{table}

\Needspace{5\baselineskip}
Under the fixed-random order, token top-$k$ spreads the budget across a median of 180.5 regions and captures 99.7\% of the deviation, but achieves only 38.6\% fidelity. Random-document ignores the deviation signal entirely, selects six whole-document spans, captures 4.8\% of the deviation, and still improves fidelity by 50.8 percentage points over token top-$k$ ($p<0.0001$). Capturing more KV deviation therefore does not guarantee higher fidelity under the evaluated fixed-random cache history: contiguous policies retain high fidelity, whereas scattered token top-$k$ does not. Neither aligning spans to document boundaries nor ranking them by KV deviation provides a statistically significant additional advantage. Document boundaries remain a convenient way to define contiguous spans.

\section{Discussion}

The same prompt can produce different answers because earlier requests change the persistent cache state that prompt encounters. This is a closed-loop serving problem: each LLM call both consumes and updates cache state, so later requests are conditioned on earlier ones. Retries, branching trajectories, concurrent sessions, and scheduler interleavings can therefore present the same prompt to a different cache state.

Because all ablation policies recompute the same number of tokens, their fidelity differences reflect which tokens they select rather than how many. Under the evaluated fixed-random order, the four contiguous policies achieve 89.4--92.8\% fidelity, compared with 38.6\% for scattered token top-$k$, even though top-$k$ captures nearly all measured KV deviation. Document-aligned recomputation and token top-$k$ also have comparable median TTFT. These results suggest that, at the same recomputation cost, recomputing contiguous regions is more effective than recomputing scattered high-deviation tokens under the evaluated fixed-random cache history.

The ablation identifies contiguity as the main factor associated with robustness under the evaluated fixed-random history. Neither aligning spans to document boundaries nor ranking them by KV deviation provides a statistically significant additional advantage. When prompts already expose document boundaries, document-aligned selection requires no learned segmenter or manually chosen chunk size, making it a convenient way to define contiguous recomputation regions. Efficient agentic serving therefore requires co-design of the recomputation policy with the persistent cache lifecycle, rather than optimizing prefill cost on isolated prompts.

% \section{Limitations}

% We do not isolate the mechanism behind history sensitivity. LMCache keys a document segment by its token sequence and does not overwrite a resident entry when the same key recurs, so across orders the KV state retained for one key may have been computed under a different surrounding prompt. If the cache fills, an entry may also be evicted and later recreated under a new prompt. Our experiments do not separate these effects. Within each order, however, the compared policies see the same request sequence, cache capacity, and budget, so the comparison uses matched serving conditions regardless of which effect dominates.

% We have not tested other models or serving stacks, nor varied the recomputation budget. Our results use Qwen3-8B with LMCache 0.5.1 and vLLM 0.23.0. We evaluate one deterministic rolling-document extraction workload; future work should test whether these findings generalize to open-ended agent tasks.

\section{Conclusion and Future Work}

We show that selective KV-cache reuse in a rolling document workload depends on persistent cache history. At a matched 5\% recomputation budget, document-aligned recomputation reduces answer variation across request orders from 69.0\% to 26.1\% relative to CacheBlend's token top-$k$. We detect no statistically significant difference under chronological order; under the other four, fidelity to full prefill improves by 34.5--52.5 percentage points at approximately the same median TTFT.

A cache-reset control detects no statistically significant difference between the methods, showing that this behavior can be missed by evaluations that run requests from a clean cache. In practice, this means that a cache-reuse method that appears reliable on isolated requests may behave differently as cache state accumulates across requests. The ablation suggests that selecting contiguous spans is the common feature associated with robustness; we detect no additional benefit from document boundaries or deviation-based span ranking.

These findings suggest that selective KV-cache reuse should be evaluated over realistic request sequences in which cache state persists, rather than only on isolated requests from a clean cache. Future work will isolate the effects of retained versus recreated cache entries and develop policies that preserve latency benefits and output fidelity as cache state accumulates.

\break
\bibliographystyle{unsrtnat}
\bibliography{references}

\end{document}